\pdfoutput=1

\documentclass[11pt]{article}

\usepackage{acl}

\usepackage{times}
\usepackage{latexsym}

\usepackage[T1]{fontenc}

\usepackage[utf8]{inputenc}

\usepackage[noend]{algpseudocode}
\usepackage{algorithmicx,algorithm}
\usepackage{bm}
\usepackage{bbm}
\usepackage{microtype}
\usepackage{booktabs}       
\usepackage{amsmath}
\usepackage{amsfonts}
\usepackage{array}
\usepackage{bm}
\usepackage{multirow}
\usepackage{subcaption}
\usepackage{hyperref}       
\usepackage{url}            
\usepackage{graphicx} 
\usepackage{utfsym}
\usepackage{xcolor}
\usepackage{inconsolata}
\usepackage{CJKutf8}
\usepackage{pythonhighlight} 
\usepackage{makecell}
\usepackage[normalem]{ulem}
\title{PyramidInfer: Pyramid KV Cache Compression \\ for High-throughput LLM Inference}

\author{Dongjie Yang\textsuperscript{\rm 1,}\thanks{\;\;Dongjie Yang and Hai Zhao are with the Department of
Computer Science and Engineering, Shanghai Jiao Tong University; Key Laboratory of Shanghai Education Commission for Intelligent Interaction and Cognitive Engineering, Shanghai Jiao Tong University; Shanghai Key Laboratory of Trusted Data Circulation and Governance in Web3.}, 
    Xiaodong Han\textsuperscript{\rm 2},
    Yan Gao\textsuperscript{\rm 2},
    Yao Hu\textsuperscript{\rm 2},
    Shilin Zhang\textsuperscript{\rm 3},
    Hai Zhao\textsuperscript{\rm 1,}\footnotemark[1]\textsuperscript{\rm ,}\thanks{\;\;Corresponding author; This paper was partially supported by Joint Research Project of Yangtze River Delta Science and Technology Innovation Community (No. 2022CSJGG1400).
    }\\
    \textsuperscript{\rm 1} Shanghai Jiao Tong University, 
    \textsuperscript{\rm 2} Xiaohongshu Inc.,\\
    \textsuperscript{\rm 3} South China University of Technology
    \\
    \textsuperscript{\rm 1}\texttt{\{djyang.tony@,zhaohai@cs.\}sjtu.edu.cn},\\
    \textsuperscript{\rm 2}\texttt{\{shuweng,yadun,xiahou\}@xiaohongshu.com}
    }

\begin{document}
\maketitle
\begin{abstract} 
Large Language Models (LLMs) have shown remarkable comprehension abilities but face challenges in GPU memory usage during inference, hindering their scalability for real-time applications like chatbots. To accelerate inference, we store computed keys and values (KV cache) in the GPU memory. Existing methods study the KV cache compression to reduce memory by pruning the pre-computed KV cache. However, they neglect the inter-layer dependency between layers and huge memory consumption in pre-computation. To explore these deficiencies, we find that the number of crucial keys and values that influence future generations decreases layer by layer and we can extract them by the consistency in attention weights. Based on the findings, we propose PyramidInfer, a method that compresses the KV cache by layer-wise retaining crucial context. PyramidInfer saves significant memory by computing fewer keys and values without sacrificing performance. Experimental results show PyramidInfer improves 2.2x throughput compared to Accelerate with over 54\% GPU memory reduction in KV cache. Our code is available in \url{https://github.com/mutonix/pyramidinfer}.

\end{abstract}

\section{Introduction}
Large Language Models (LLMs) \citep{openai2023gpt4, anthropic2023claude, jiang2023mistral} like GPT4 have demonstrated the unprecedented ability of remarkable comprehension in human languages. However, these large models meet up with a substantial challenge of immense GPU memory usage in the inference, due to the model and computational complexity. This hinders deploying LLMs at scale to meet the thousands of demands for chatting with chatbots. 

\begin{figure}[ht]
    \centering
    \includegraphics[width=0.9\linewidth]{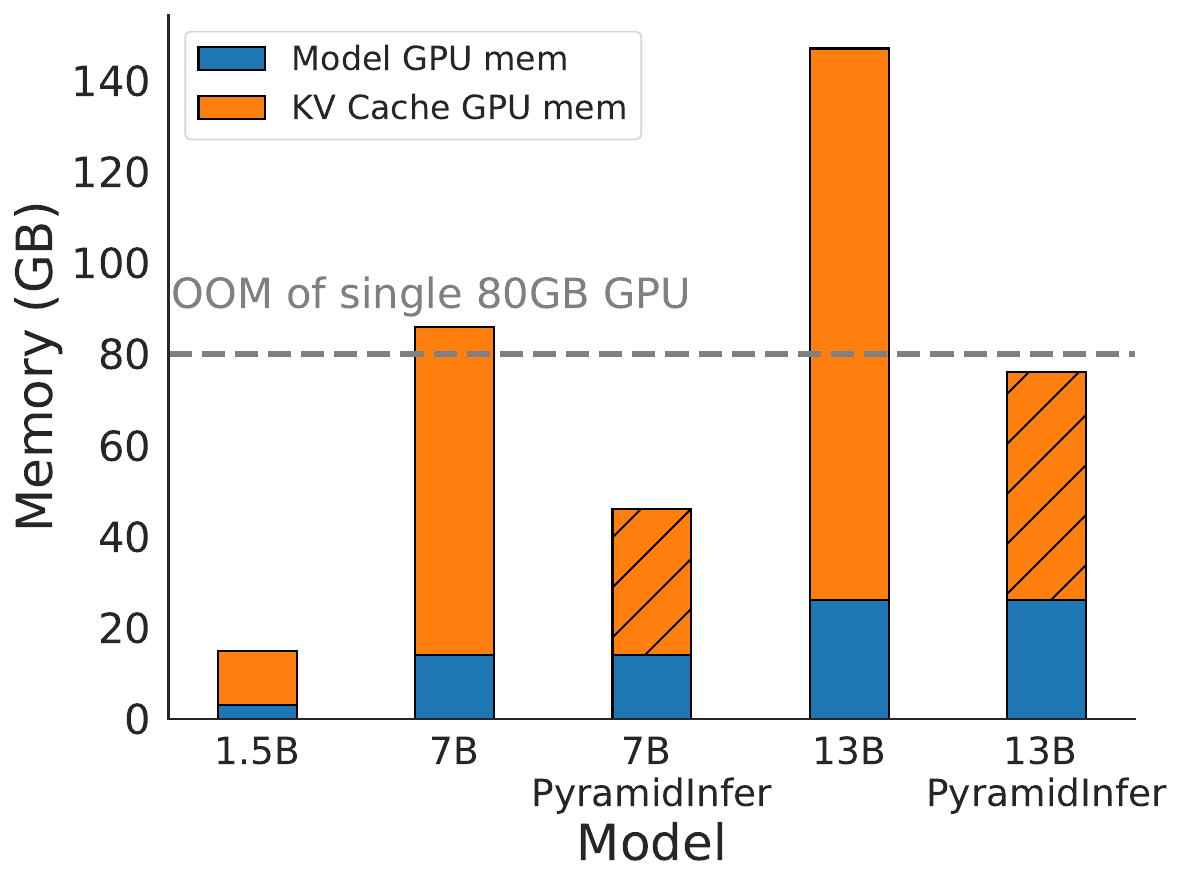}
    \caption{Inference in the prefill phase: all models of different sizes have the prompts of 64 $\times$ 2k. LLM consumes huge GPU memory in the KV cache compared to the small model. PyramidInfer can reduce over 54\% GPU memory usage in the KV cache while having more than 2x throughput.}
    \label{fig:memory}
\end{figure}

Different from training, models in the inference do not need to record the optimizer states, activations, or gradients. As LLMs are mostly Transformer-based auto-regressive models, the GPU memory usage mainly consists of two parts: model parameters and KV cache. KV cache presents the keys and values previously computed in the attention. We store the KV cache in the GPU memory and reuse it in future generations to avoid re-computation. The KV cache mechanism has been widely used to improve the inference speed \citep{touvron2023llama, zhang2022opt}. However, the KV cache consumes huge GPU memory, especially for LLMs. For example, in Figure \ref{fig:memory}, for a model with 7 billion parameters, the parameters only consume 14 GB of memory but the KV cache requires around 72 GB. The KV cache has the potential to consume memory several times the size of the model. It demonstrates a great challenge that the throughput of LLM inference is constrained by how much data (KV cache) we can put in the GPU besides the model.

\begin{figure*}[ht]
    \centering
    \includegraphics[width=0.95\linewidth]{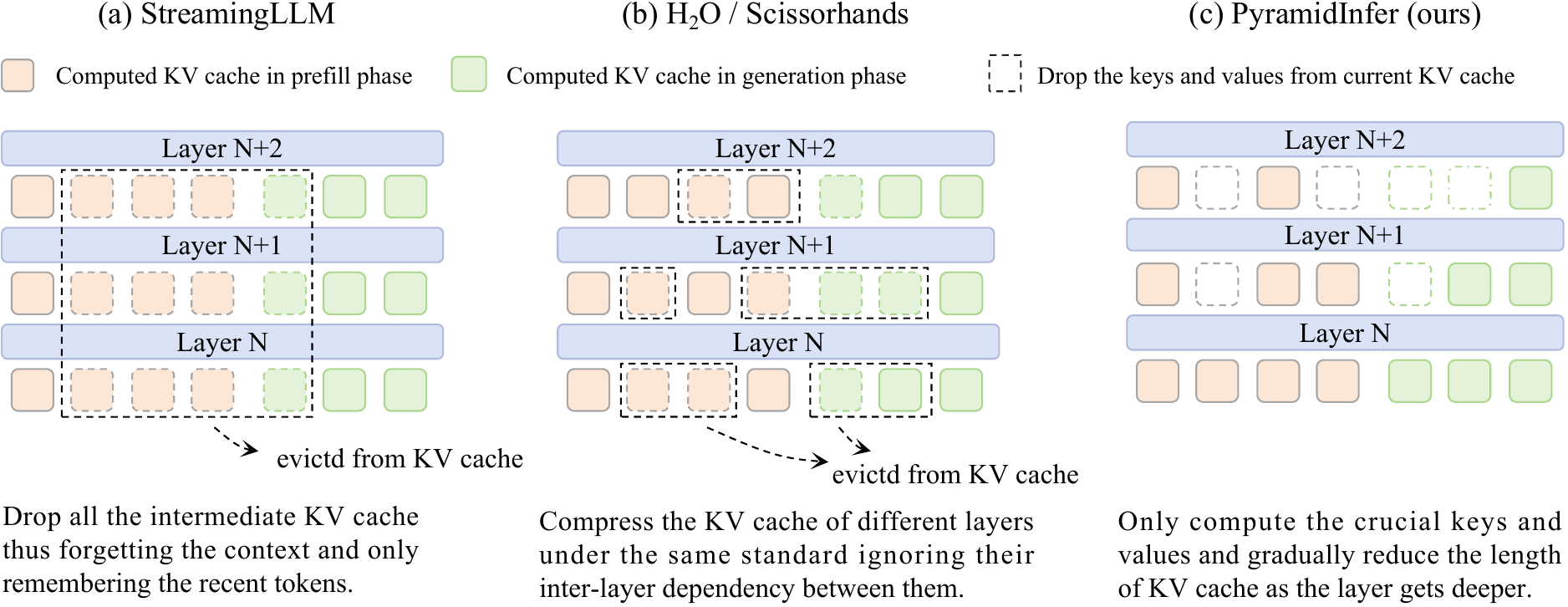}
    \caption{Comparison between PyramidInfer and other methods: (a) StreamingLLM only reserves the first and recent tokens thus losing memorization of the previous context. (b) H$_2$O/Scissorhands compress the KV cache without difference for all the layers. They suffer great information loss by compressing too much in the shallow layers. (c) Different from the above methods that can only compress after the KV cache has been computed, PyramidInfer can compress the KV cache in the prefill phase. PyramidInfer only computes crucial keys and values to do inference thus reducing more GPU memory and bringing higher throughput.
    }
    \label{fig:comparison}
\end{figure*}

We break down LLM inference into two phases: prefill phase and generation phase \citep{brown2020language, radford2019language}. In the prefill phase, the prompt is computed in parallel to generate the first token, and the initial KV cache is pre-filled. In the generation phase, the model decodes the next token one by one and appends the keys and values of the newly decoded token to the old KV cache. Recent studies \cite{zhang2023h2o, liu2023scissorhands, ge2023model} compress the KV cache to reduce GPU memory usage. However, as shown in Figure \ref{fig:comparison}, they all only reduce the KV cache that has been already computed rather than reducing the KV cache to be computed. They have to prefill the initial KV cache before they can start to compress, which neglects the great GPU memory consumption of computing the initial KV cache, especially for longer prompts and larger models. If the model can not process the prompt in the prefill phase, these methods are no longer applicable as their compression starts in the generation phase. In this paper, we focus on how to further compress the KV cache in the prefill phase besides the generation phase. We give out our findings and then propose our method PyramidInfer inspired by these findings.

During the training, all input tokens predict the tokens next to themselves in an one-to-one teacher-forcing way \citep{lamb2016professor}. During the inference, the tokens except for the last token no longer need to predict the next tokens but they still record this redundant information in keys and values. We call this \textbf{Inference Context Redundancy} (ICR) hypothesis. It inspires us to compress the KV cache by only computing the keys and values that record the context information.

Another challenge arises as the initial KV cache is reused multiple times for generating future tokens, necessitating careful retention of context information during compression. Inspired by the work \citep{liu2023scissorhands}, we further explore what parts of the KV cache are always crucial for future generations. We observe that queries of recent tokens closer to the last token are more consistent in attending to the same context keys and values, denoted as the Pivotal Context (PvC). We call this phenomenon as \textbf{Recent Attention Consistency} (RAC). The consistency of attention weights in recent tokens indicates that we can leverage it as the oracle to select the crucial KV cache for future generations in advance. 

Based on our observations, we propose the PyramidInfer, an effective method of reducing the KV cache both in the prefill and generation phase by layer-wise selecting the PvCs. In PyramidInfer, the PvCs are gradually reduced as the layers get deeper where the KV cache is like a pyramid. We showcase the capability of PyramidInfer on a wide range of tasks using OpenCompass \citep{2023opencompass} on models of different types and sizes. The results show that PyramidInfer has higher throughput than the full cache method Accelerate and Deepspeed by 2.2x and 1.4x, KV cache compression method H$_2$O by 2.4x with over 54\% less GPU memory in KV cache.

\section{Related Work}
Due to the increasing demands for chatting with chatbots, efficient strategies are required to process thousands of queries to maximize the throughput. The fundamental way to improve the throughput is to put more data (larger batch) into the GPU memory to utilize the GPU parallelism better. 

\paragraph{Inference Parallelism}
One way is to enlarge the GPU memory. We can borrow the techniques used in training to accelerate the inference, e.g., pipeline parallelism \citep{huang2019gpipe}, KV cache offload \citep{sheng2023flexgen}, etc. These methods leverage multiple GPUs or even RAM to make up bigger space for input data. 

\paragraph{KV Cache Reduction}
However, if we have limited GPU memory, another way is to reduce the KV cache. For optimization in the CUDA, FlashAttention 2 \citep{dao2023flashattention2} reduces the number of reads/writes between GPU HBM and GPU on-chip SRAM. PagedAttention \citep{kwon2023efficient} borrows the virtual memory techniques to achieve near-zero waste in KV cache memory.

Besides CUDA methods, we can optimize the KV cache from the model itself. From Figure \ref{fig:comparison}, StreamingLLM \citep{xiao2023efficient} reserves the recent context to enable unlimited input by sacrificing memorization of the history. Other methods like H$_2$O \cite{zhang2023h2o} and Scissorhands \citep{liu2023scissorhands} leverage the attention to compress the KV cache. However, they treat the compression of different layers as the same thing and can not compress in the prefill phase. Our method PyramidInfer takes the difference in layers into account and realizes the compression in both the prefill and generation phases, thus better reducing the KV cache while maintaining the generation quality.

\begin{figure*}[ht]
    \centering
    \includegraphics[width=0.92\linewidth]{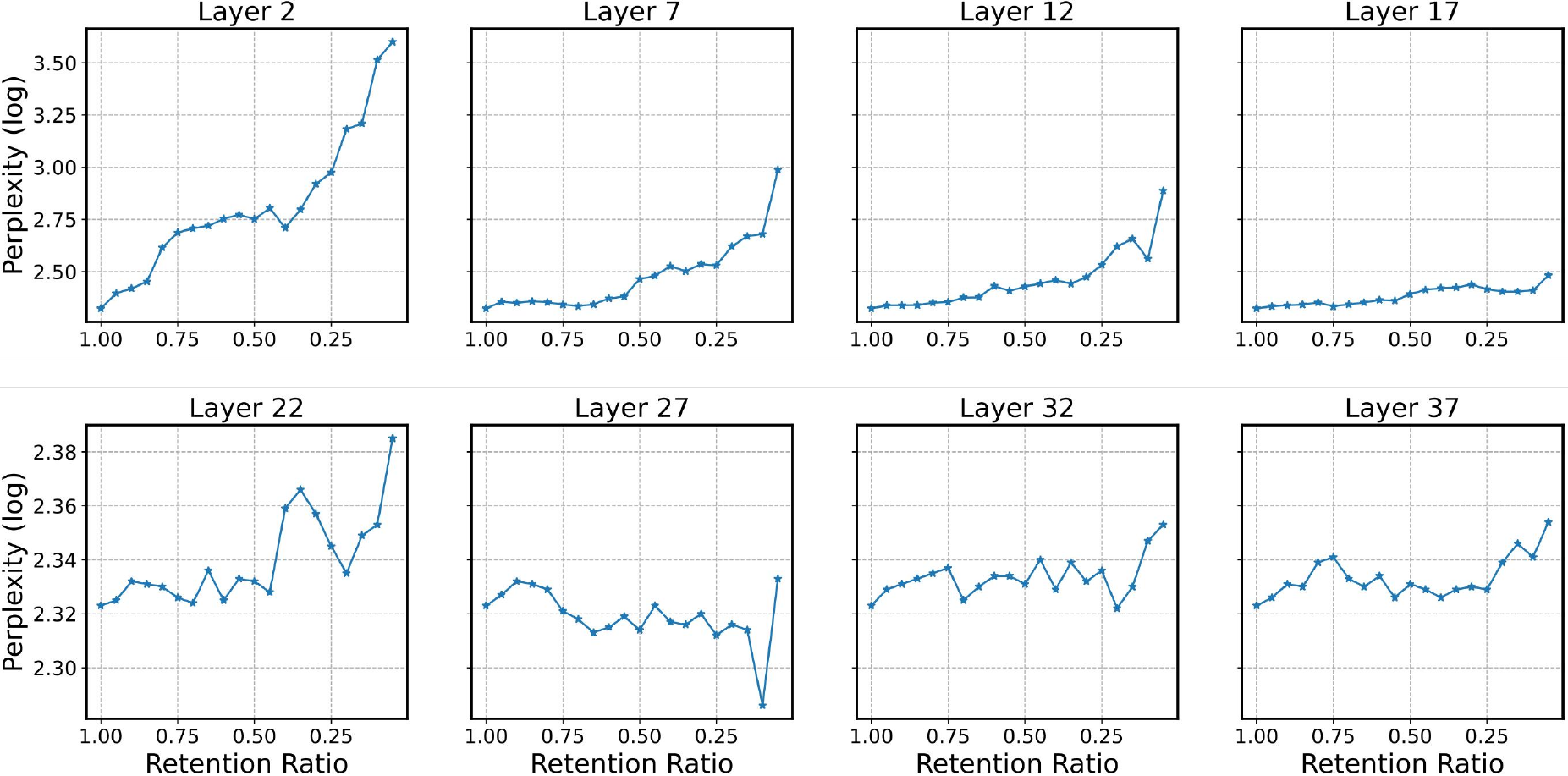}
    \caption{For each layer, we reserve the keys and values with top-$p$ attention weights (PvC) while other layers maintain the full length. We calculate the average perplexity across different retention ratios $p$.}
    \label{fig:cpr}
\end{figure*}

\begin{figure}[h]
    \centering
    \includegraphics[width=0.75\linewidth]{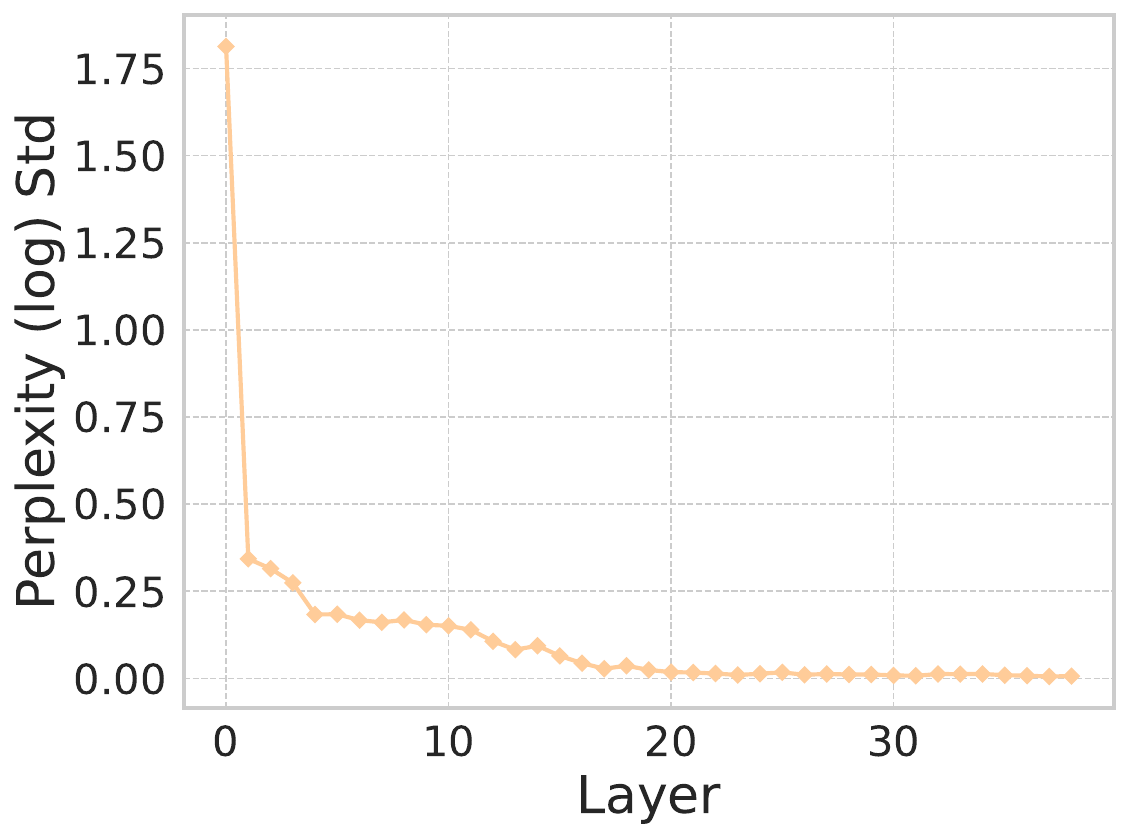}
    \caption{The perplexity standard deviations when only PvCs are reserved at each layer.}
    \label{fig:cpr_std}
\end{figure}

\section{Observation and Insight}
\label{sec:observation}
We verify the hypotheses of Inference Context Redundancy and Recent Attention Consistency, which inspire us to design the method \textbf{PyramidInfer}.

\subsection{Inference Context Redundancy}
Different from teacher-forcing in the training, only the last token has to predict the next token in the inference. We suppose there exist keys and values of the context that record the redundant information to predict the next token in the training but are not useful for inference. We call this the Inference Context Redundancy (ICR) hypothesis. 

\subsubsection{Pivotal Context} To verify the hypothesis, we design an experiment based on 40-layer LLaMA 2-13B to find out if this redundancy exists in the KV cache. In this experiment, we only reserve a proportion of keys and values of certain layers while other layers remain fixed and see how the perplexity of model output will change. This selected proportion consists of the important keys and values with the top-$p$ attention weights, denoted as the Pivotal Context (PvC).

As shown in Figure \ref{fig:cpr}, we show that, for most of the layers, as the retention ratio of PvC decreases, the perplexity of the output will increase. However, as the layer becomes deeper (larger index), we find that the influence of shorter PvC tends to be smaller. For example, after Layer 27, the perplexity remains stable even with 80\% keys and values are evicted. In Figure \ref{fig:cpr_std}, we compute the standard deviations across the retention ratios of all the layers and observe they obey a power law distribution. It indicates most of the keys and values should be retained as the layers are shallow and the redundancy in the KV cache sharply increases as the layers become deeper. This growing redundancy guides us to minimize the KV cache while maximizing the performance.

\subsubsection{Discussion}
\paragraph{How does the model gather information to predict the next token?}
Generating the next token can be considered as a process that the last token gathers the information from the context based on the attention weights. In Figure \ref{fig:cpr}, we observe from the view of the last token. In the shallow layer, the information in the context is distributed in most of the tokens in the context. As the layer goes deeper, only limited keys and values contribute to the next token prediction. 

The inference process differs from training because all the input tokens predict the next tokens. At this time, keys and values store two kinds of information: 1) the information to predict what the token is next to it; 2) the context information for future tokens to leverage. So far, we have verified that PvCs are the crucial keys and values that are useful for inference. On the other hand, we want to verify the non-PvC that may play a more important role in teacher-forcing prediction instead of being the context. As non-PvCs are trivial in PyramidInfer, we discuss it in the Appendix \ref{app:icr}. 

\begin{figure*}[ht]
    \centering
    \begin{subfigure}{0.9\textwidth}
    \includegraphics[width=1.0\linewidth]{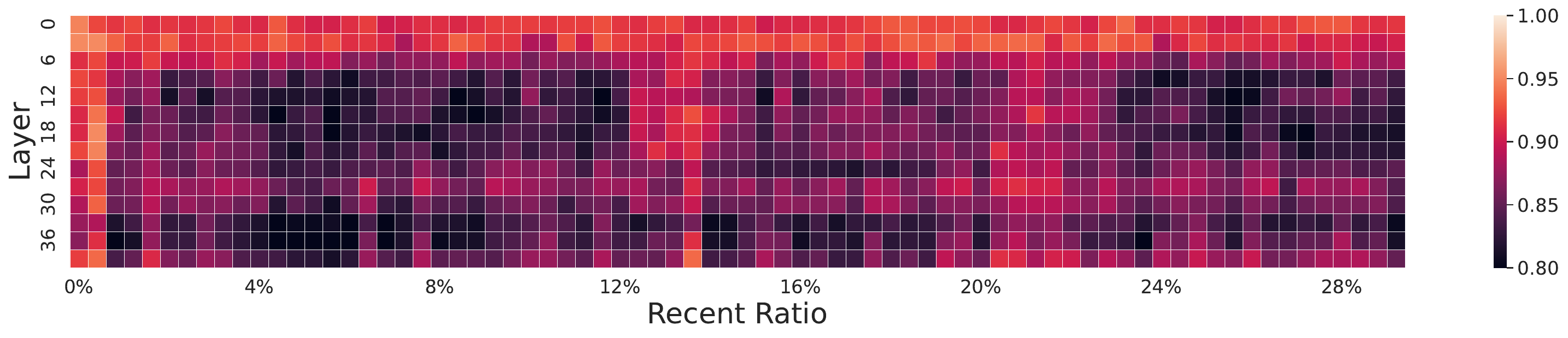}
    \caption{Separate PvC overlap ratios of recent tokens.}
    \label{fig:rac_before}
    \end{subfigure}
    \\
    \begin{subfigure}{0.9\textwidth}
    \includegraphics[width=1.0\linewidth]{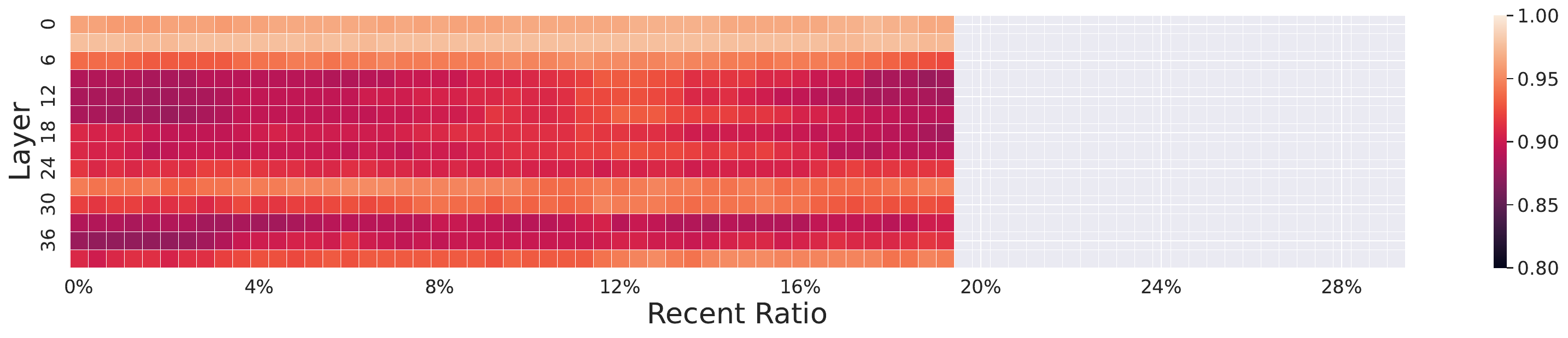}
    \caption{Ensemble PvC overlap ratios of recent tokens.}
    \label{fig:rac_after}
    \end{subfigure}
    \caption{PvC overlap ratio heatmap.}
\end{figure*}

\subsection{Recent Attention Consistency}
In the verification of ICR, we use the attention weights to find PvCs. However, in an attention layer, there are several attention weights for one token $x_i$ as every subsequent token $x_{t > i}$ will attend to it. Which attention weights should we choose as the metric to find PvCs? Intuitively, the optimal weights must be from the last token $x_n$. However, the PvCs selected by these weights are suitable for predicting $x_{n+1}$ but not always suitable for future tokens $x_{t > n+1}$. Our goal is to find if there exists shared PvCs that can be used as a general oracle to predict several future tokens $x_{t>n+1}$ besides the last token $x_{n+1}$.

\subsubsection{PvC Consistency}
We convert this goal to finding if there exist keys and values that are frequently attended by subsequent tokens. First of all, we define a relative distance of how far the context token $x_i$ is relative to the last token $x_n$, which is called the Recent Ratio $d = (n - i)/n \times 100\%$.  We divide the input sequence into two parts where we denote the tokens with $0 < d < 30\%$ as the recent sequence $S_r$ and $d \geq 30\%$ as the context sequence $S_c$. We only compute the attention weights of $S_r$ to $S_c$ to check if there are tokens in the $S_c$ that are always attended by the tokens in the $S_r$. For each token in $S_r$ of each layer, we select the keys and values with top-80\% attention weights as their PvCs. We set the keys and values with top-80\% attention weights of the last token ($d=0$) as the PvC selection baseline. 

After the setup, we want to measure how much the overlap will be that the PvCs of recent tokens are consistent with the PvC of the last token. If there is overlap, we can infer the intersection should be the shared PvC where many subsequent tokens are consistently interested. Thus for each layer $l$, we calculate the overlap ratio $C$ of PvCs as follows:
\begin{equation}
\begin{gathered}
    C_{l,i} = 
    \frac{|\{x | x \in \textbf{PvC}_{l,i}\} \cap  \{x | x \in \textbf{PvC}_{l,last}\}|}{|\{x | x \in \textbf{PvC}_{l,last}\}|}.
\end{gathered}
\end{equation}

From the results in Figure \ref{fig:rac_before}, the recent tokens in $S_r$ have an average 86\% overlap with the PvC selected by the last token. It indicates there exists shared PvCs that are always interested in by the subsequent tokens. However, it is not enough to be the oracle to predict future tokens. For example, if we want to predict the $x_{n+1}$ token using only the PvC extracted from the token with $d=25\%$, we only have about 83\% PvC contributes to the prediction, which suffers a great context information loss. 

Fortunately, the PvC selections from recent tokens have high consistency and we can integrate multiple tokens to select the shared ones. In Figure \ref{fig:rac_after}, we integrate the attention weights by averaging weights of subsequent $[d, d + 10\%]$ tokens as the ensemble weights of the token with $d$.
We select the keys and values with top-80\% ensemble weights as PvCs. We observe that the average PvC overlap ratios increase by a large margin to approximately 93\%. The overlap ratios have hardly any drop with $d = 20\%$, which indicates we can leverage the PvCs selected from ensemble tokens with $d = 20\%$ as an oracle to predict the $x_{n+1}$ which is 20\% ahead.

\begin{figure*}[t]
\begin{minipage}{.44\textwidth}
	\centering
	\includegraphics[width=1.0\textwidth]{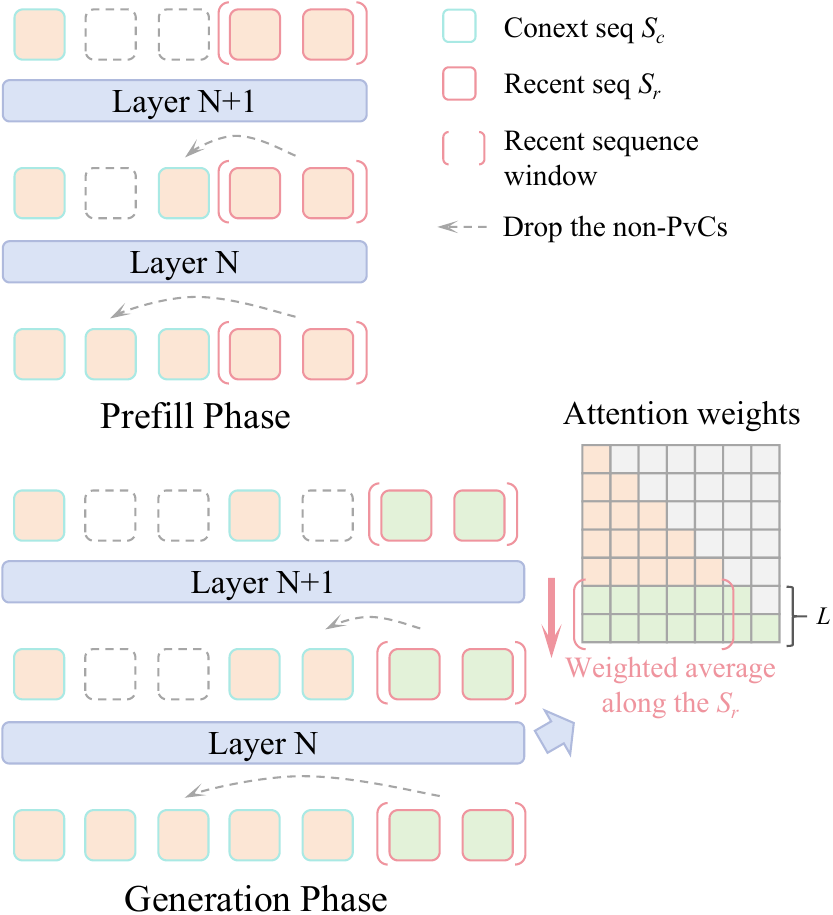}
	\caption{The overview of the PyramidInfer.}
	\label{fig:overview}
\end{minipage}
\hfill
\begin{minipage}{.55\textwidth}
\begin{algorithm}[H]
\caption{One forward pass in PyramidInfer}
\label{al:PyramidInfer}
\hspace*{0.02in} {\bf Input:} 
KV cache $KV$, recent window length $L$, min PvC length $\mathbf{N} = \{N_0, \dots, N_l, \dots\}$
\\
\hspace*{0.02in} {\bf Output:} updated KV cache $KV$
\begin{algorithmic}
\For {layer $l$ $\in$ layers}
    \If {$KV$ is not None} 
        \State $KV = \mathrm{cat}([\mathbf{PvC}_{past}, KV])$
    \EndIf
    \State $\mathcal{A} \leftarrow \mathrm{compute\ attention\ weights\ of\ } KV$
    \State $\mathcal{A}_{e} \leftarrow \mathrm{weighted\_avg}(\mathcal{A}[-L:, :-L], \mathrm{dim=-2})$
    \If {$\mathrm{len}(KV) > N_l$}
        \State $\mathrm{TopP\_index} \leftarrow \mathrm{TopP}(\mathcal{A}_e,\ p=p)$
        \State $\mathbf{PvC} \leftarrow \mathrm{Gather}(KV, \mathrm{index=TopP\_index})$
    \EndIf
    \State $KV \leftarrow \mathbf{PvC}$
    \State Reduce $p$ by multiplying a decay ratio
\EndFor
\Return $KV$
\end{algorithmic}
\end{algorithm}
\end{minipage}
\end{figure*}

\subsubsection{Discussion}
\label{sec:rac_discuss}
\paragraph{Why do the deeper layers tend to have lower PvC overlap ratios?}
If we check overlap ratios along the layer axis, we find that only shallow layers have relatively high ratios. It is because in deeper layers there is context redundancy: Only a small number of keys and values have high weights that are always selected as PvCs; The others have similar low weights so they are not always selected, which results in lower overlap ratios. This phenomenon is consistent with the power law distribution observed in ICR, which is further discussed later.

\paragraph{Context information is mostly stored in the shared PvCs.}
In Figure \ref{fig:rac_after}, the consistent PvC overlap ratios from small $d$ to large $d$ show that wherever recent tokens are, they only leverage nearly the same number of keys and values in the context. These keys and values, also known as shared PvCs, store most of the context information.

\section{Layer-wise PvC Selection}
\label{sec:method}
Based on the observations, we design the PyramidInfer, a method to highly increase the inference throughput by layer-wise selecting the PvCs to compress the KV cache for each layer.

\subsection{Method}
As shown in Figure \ref{fig:comparison}, PyramidInfer can not only reduce the KV cache in the generation phase but also in the prefill phase without computing the complete keys and values of the prompt for all the layers. Following the inference process, we introduce the PyramidInfer in the prefill phase and generation phase separately and see how PyramidInfer can save lots of GPU memory by carefully selecting the PvCs. 

\paragraph{Prefill Phase}
In the prefill phase, we have to process the prompt to prefill the initial KV cache. Different from the common inference process that reserves all keys and values of the prompt, PyramidInfer only reserves the PvCs of each layer as the initial KV cache.

Similarly, we divide the input sequence into recent sequence $S_r$ and context sequence $S_c$. As shown in Algorithm \ref{al:PyramidInfer}, based on the RAC, we first calculate the ensemble attention weights by weightedly averaging the attention weights of $S_r$. We assign larger weights for more recent tokens to enlarge their impact on PvC selection. Based on the ensemble attention weights, we layer-wise select the keys and values with top-$p$ weights as the PvC. According to the conclusion of ICR, the increment of redundancy obeys the power law distribution. We choose a larger $p$ to retain more tokens in the $S_c$ for not to lose the semantics in the shallow layers. Then we gradually decrease the $p$ to reduce the length of PvCs in deeper layers. Therefore, the PvCs of the deeper layers are shorter and the KV cache becomes a "pyramid".

The layer-wise PvC selection saves much more GPU memory than other methods computing the whole prompt in the prefill phase. Besides the prefill phase, PyramidInfer continues to boost efficiency in the generation phase because LLMs only need to reuse a smaller initial KV cache.

\paragraph{Generation Phase}
As we have reserved the initial PvCs as the KV cache, what we should do in the generation phase is to update these PvCs according to the new recent tokens. As shown in Figure \ref{fig:overview}, we maintain a sliding recent window to update the newly generated token to be new recent tokens. Based on the new $S_r$, we update the PvCs of the KV cache where the operation is the same as the prefill phase. By controlling the length of the PvC of each layer, we can easily tune the compression ratio and even support unlimited input like StreamingLLM by maintaining a fixed number of PvCs in the KV cache.
\begin{figure*}[ht]
    \centering
    \includegraphics[width=1.0\linewidth]{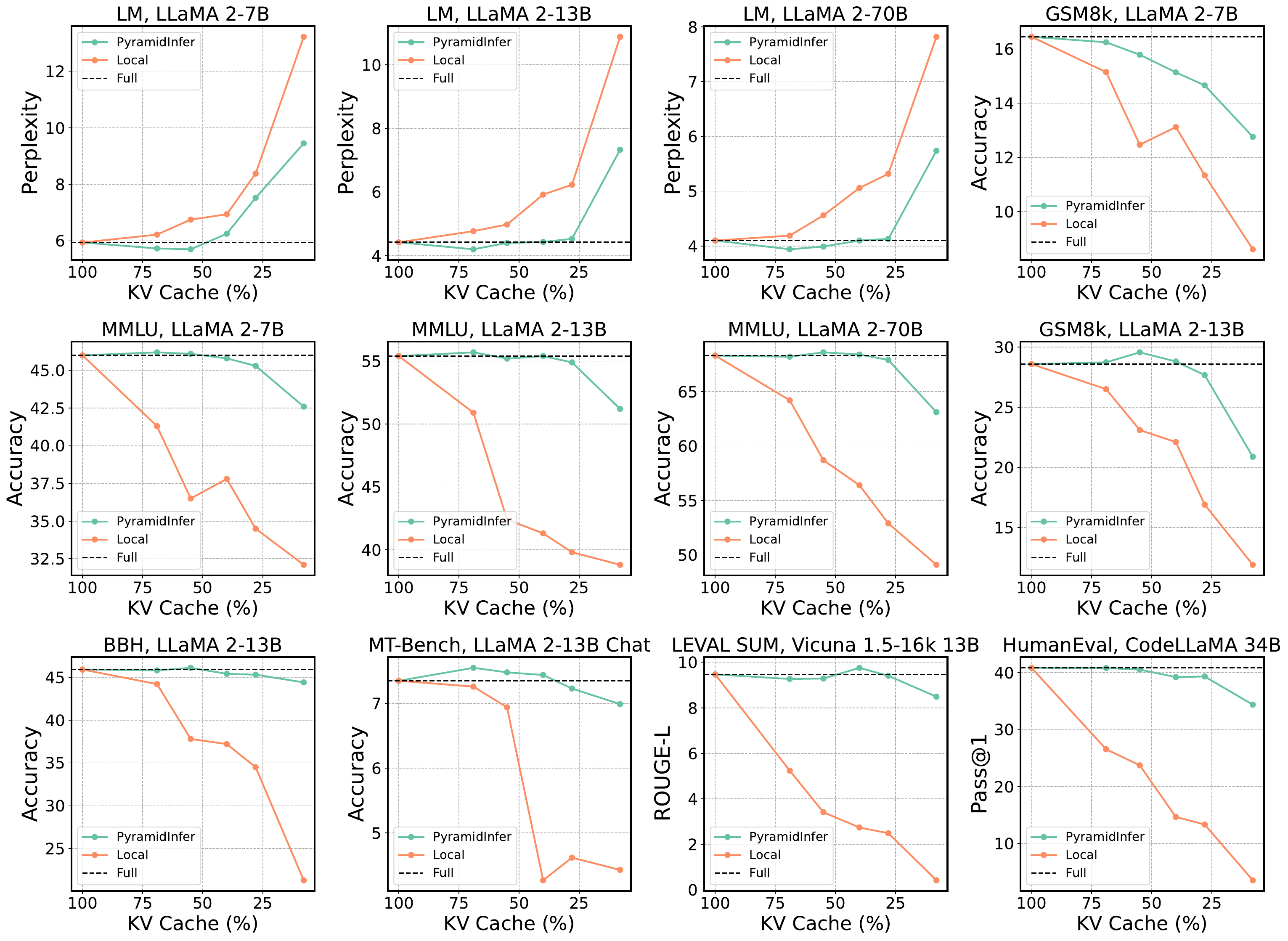}
    \caption{Benchmark results of comparison between models with full cache, "local" strategy, and PyramidInfer.}
    \label{fig:result}
\end{figure*}

\section{Evaluation}

\begin{figure*}[t]
\makeatletter\def\@captype{table}\makeatother
\begin{minipage}{.62\textwidth}
\centering
\setlength{\tabcolsep}{5.pt}
\renewcommand\arraystretch{1.15}
\caption{The evaluation of inference methods using an A100 80GB GPU on LLaMA 2-13B and 70B. \textbf{Length}: prefill length + generation length. \textbf{Bsz}: batch size. \textbf{KV mem.}: GPU memory usage (GB) of the KV cache. \textbf{Thr.}: throughput (token/s)}
\label{tab:memory_result}
 \small
{
\begin{tabular}{l|cc|c|cc}
\toprule
{\textbf{Model}} & \textbf{Bsz}  & \textbf{Length} & \textbf{Method} & \textbf{KV Mem.} & \textbf{Thr.}\\
\midrule
\multirow{4}{*}{13B} &  \multirow{4}{*}{32} &\multirow{4}{*}{512+256} & Accelerate &  24.2 (100\%) & 621 (1.0x)\\
 &  & & Deepspeed & 24.2 (100\%) & 934 (1.5x)\\
 &  & & H$_2$O & 21.6 (89.2\%)  & 584 (0.9x) \\
\cline{4-6}
 & & & PyramidInfer  & \textbf{11.0 (45.4\%)} & \textbf{1389 (2.2x)} \\
\midrule
\multirow{2}{*}{70B} &  \multirow{2}{*}{8} & \multirow{2}{*}{256+128} & \makecell{Accelerate/\\Deepspeed/H$_2$O} & OOM & - \\
\cline{4-6}
 &  &  & PyramidInfer & 4.2 & 20\\
\bottomrule
\end{tabular}
}
\end{minipage}
\hfill
\makeatletter\def\@captype{figure}\makeatother
\begin{minipage}{.36\textwidth}
\centering
\includegraphics[width=1.0\textwidth]{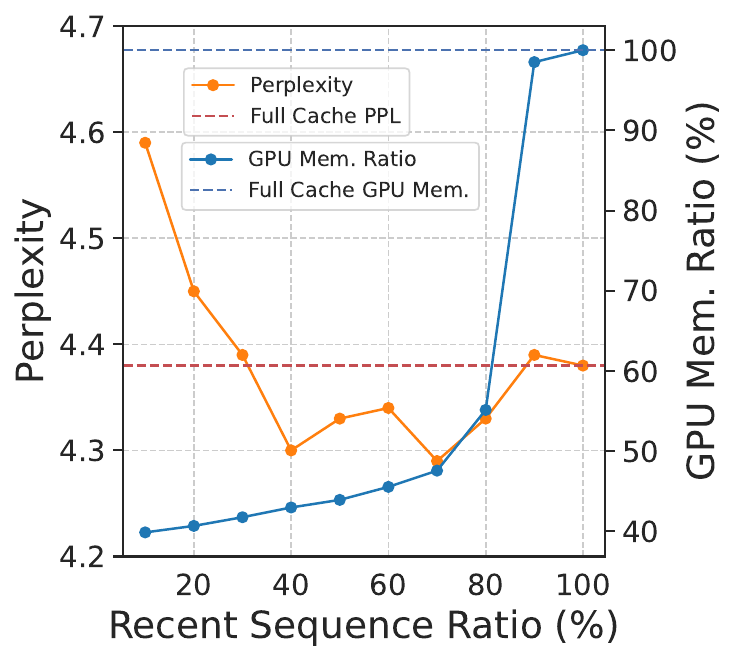}
\caption{$S_r$ ratio ablation study.}
\label{fig:recentratio}
\end{minipage}
\end{figure*}

\begin{table*}[ht]

\end{table*}

\subsection{Basic Evaluation}
We evaluate PyramidInfer on various tasks and models to showcase that PyramidInfer can largely reduce the GPU memory and increase the throughput while maintaining the generation quality.
\paragraph{Experimental Setup}
We choose four kinds of scenarios: 1) Language modeling: we measure the perplexity on wikitext-v2 \citep{merity2016pointer}. 2) LLM benchmarks: we evaluate on MMLU \citep{hendrycks2021measuring} and BBH \citep{srivastava2022beyond} for language understanding, GSM8K \citep{cobbe2021training} for mathematical reasoning, HumanEval \citep{chen2021evaluating} for coding. 3) Conversation: We evaluate on MT-Bench \citep{zheng2023judging} to see how PyramidInfer can handle multi-turn conversation. 4) Long context: we evaluate on long text summarization of the LEval \citep{an2023leval} to see if PyramidInfer can maintain the quality while accepting longer input. We evaluate these tasks on LLaMA 2 \citep{touvron2023llama}, LLaMA 2-Chat, Vicuna 1.5-16k \citep{zheng2023judging} and CodeLLaMA \citep{rozière2023code} with different sizes (7B, 13B, 34B and 70B) \footnote{We quantize the 34B and 70B models to INT8 data type to reduce the computational cost.}.  We set the full KV cache method as the baseline. Besides that, we also include the "local" strategy as another baseline that reserves only the recent KV cache.

In addition, we showcase how much PyramidInfer can save GPU memory and improve the throughput. We compare the efficiency of PyramidInfer with other full cache methods, including Accelerate \citep{huggingface2021accelerate}, Deepspeed\footnote{\url{https://github.com/microsoft/DeepSpeedExamples/tree/master/inference}} \citep{aminabadi2022deepspeed}. We also select H$_2$O\footnote{\url{https://github.com/FMInference/H2O}}  \citep{zhang2023h2o}, a KV cache compression method, as another baseline. It is noted that PyramidInfer is orthogonal to the non-KV-compression methods like Deepspeed to improve efficiency further.

\paragraph{Benchmark Result}
In Figure \ref{fig:result}, we evaluate the LLMs with different compression ratios. We show that PyramidInfer maintains the generation quality with much less GPU memory compared with the full cache baseline. PyramidInfer also outperforms the "local" strategy with a large gap across different types and sizes of models and tasks.

In the LEval that tests the long context ability, we show that the "local" strategy that is similar to the technique used in StreamingLLM causes a huge decline in memorization of history. PyramidInfer can accept longer input with less GPU memory without sacrificing too much performance.

\paragraph{Efficiency Result}
In Table \ref{tab:memory_result}, we fix the input length and the batch size. For LLaMA 2-13B, PyramidInfer showcases 2.24x throughput than full cache using Accelerate with 54.6\% less GPU memory in the KV cache. For LLaMA 2-70B, PyramidInfer can still generate in the prefill phase compared to other methods. Existing KV cache compression methods like H$_2$O can not even process the prompt and strike the OOM before the start of compression.

In Table \ref{tab:throughput_result}, we exhaust the memory of an 80GB A100 GPU to test the maximum throughput by maximizing the batch sizes. PyramidInfer enables more than 2x batch size than others and has higher throughput than full cache methods Accelerate and Deepspeed by 2.8x and 1.7x, KV cache compression method H$_2$O by 2.1x. PyramidInfer can also be utilized to enhance Deepspeed by increasing the throughput by 1.9x.

\begin{table}[ht]
\centering
\setlength{\tabcolsep}{5.pt}
\renewcommand\arraystretch{1.15}
\caption{We exhaust the memory of an A100 80GB GPU to find out the maximum throughput of these methods on LLaMA 2-13B. We set the input length to 512+256. \textbf{Lat.}: latency to generate one token (ms/token).}
\label{tab:throughput_result}
 \small
{
\begin{tabular}{l|c|ccc}
\toprule
\textbf{Method} & \textbf{Max Bsz} & \textbf{Lat.} & \textbf{Thr.} \\
\midrule
Accelerate & 42 & 1.72 (100\%)  & 581 (1.0x) \\
Deepspeed & 40 & 1.03 (59.8\%)  & 972 (1.6x)\\
H$_2$O & 48 & 1.39 (80.8\%) & 769 (1.3x) \\
\cline{1-4}
PyramidInfer & 88 & 0.59 (34.3\%) & 1678 (2.8x) \\
\makecell{PyramidInfer\\+Deepspeed} & 86 & \textbf{0.53 (30.8\%)} & \textbf{1887 (3.2x)} \\
\bottomrule
\end{tabular}
}
\end{table}

\subsection{Ablation Study}
We conduct the ablation studies using the LLaMA 2-13B model to explore the PyramidInfer by answering the following questions: 1) Which way should we choose to gradually reduce the PvC length as the layer becomes deeper without sacrificing too much performance? 2) What proportion of the input should we partition as the recent sequence $S_r$? 

\begin{table}[ht]
\centering
\setlength{\tabcolsep}{4.5pt}
\caption{PvC length decay ablation study.}
\label{tab:decay_result}
 \small
{
\begin{tabular}{l|ccc}
\toprule
{\textbf{Strategy}} & \textbf{PPL} & \textbf{GSM8K} & \textbf{MMLU} \\
\midrule
Reduce more & 4.93 & 26.82 & 53.1 \\
Reduce uniformly & 4.55 & 28.32 & 54.8\\
Reduce less (PyramidInfer) & 4.20 & 29.56 & 55.7 \\
Reduce None (Full cache) & \textbf{4.42} & \textbf{28.58} & 55.4 \\
\bottomrule
\end{tabular}
}
\end{table}

\paragraph{PvC Length Decay}
Based on ICR, we gradually reduce the length of PvCs for each layer as the layer becomes deeper to maximize efficiency. However, excessive reduction of PvC length in shallow layers may lead to the loss of context information. We try to find out which way is the best to reduce the PvC length. Under the same compression ratio of 60\%, we compare three patterns: 1) reduce more PvC length in shallow layers but less in the deeper layers (reduce 15\% cache in the first 50\% layers). 2) uniformly reduce the PvC length (reduce 10\% cache in the first 50\% layers); 3) obey the power law pattern based on ICR to reduce less at first (reduce 7\% cache in the first 50\% layers).

The result in Table \ref{tab:decay_result} demonstrates that following the power law pattern is the best way to reduce the PvC length and even slightly improve performance on downstream tasks.

\paragraph{Recent Sequence Ratio}
In PyramidInfer, we select the recent tokens of the input as the recent sequence $S_r$. The $S_r$ is not only leveraged as the context but also the criteria to select the PvC from the context sequence $S_c$. If the $S_r$ ratio increases, $S_c$ will be shorter thus fewer tokens in $S_c$ will be compressed. Therefore, we need to find a balance to decide how large the $S_r$ ratio should be.

In Figure \ref{fig:recentratio}, we set the GPU memory usage of the KV cache of the full cache method as the 100\% baseline and test how the perplexity will change with different $S_r$ ratios. As the $S_r$ ratio increases, we observe a decline in the GPU memory usage but a trough in the perplexity at 40-60\% $S_r$ ratio. Thus we can choose 40\% as a trade-off between performance and GPU memory usage.

\section{Conclusion}
We alleviate the difficulty of deploying LLMs at scale by introducing PyramidInfer, a novel method that efficiently compresses the KV cache during both prefill and generation phases. Inspired by ICR and RAC, PyramidInfer significantly reduces GPU memory usage without compromising model performance. Experimental results present PyramidInfer is a promising solution for optimizing LLM deployment in resource-constrained environments.

\section*{Limitations}
Despite the effective strategy to reduce the keys and values to be computed by selecting the PvCs, PyramidInfer has to bring in additional computation so that it has limited speedup with a small batch size, as discussed in Appendix \ref{app:cost}.

Besides that, we are the pioneers in compressing the KV cache in the prefill phase, which is an area not fully explored. PyramidInfer is not a method to compress the KV cache losslessly in the prefill stage and more effective methods can be explored in future works.

\bibliography{custom}
\appendix

\clearpage
\section{Extended Experiments and Details}
\subsection{Additional Computational Cost in PyramidInfer}
\label{app:cost}
\begin{figure}[ht]
    \centering
    \includegraphics[width=0.8\linewidth]{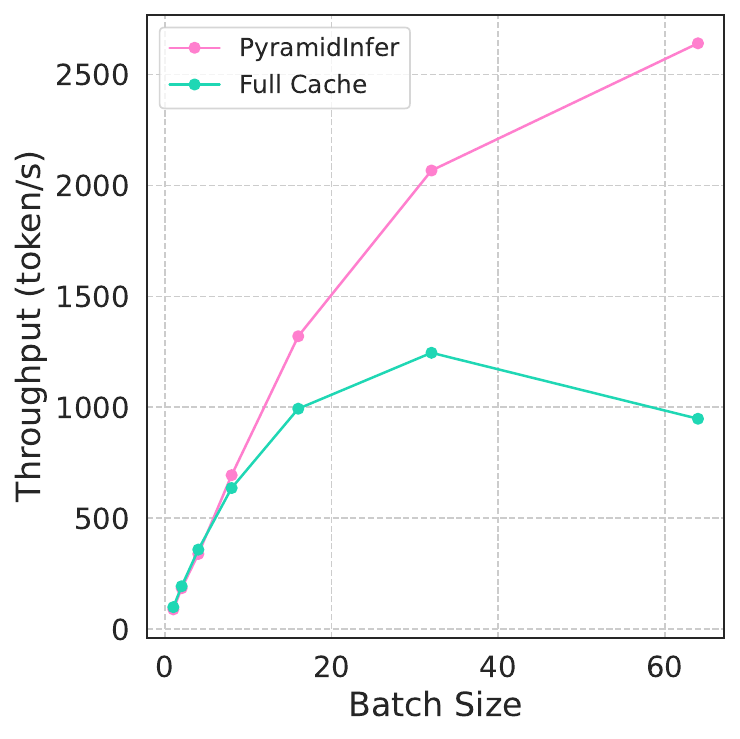}
    \caption{Comparison between PyramidInfer and full cache baseline with different batch sizes on the LLaMA 2-7B model with input length of 512+256.}
    \label{fig:cost}
\end{figure}

In Section \ref{sec:method}, we introduce how PyramidInfer improves the inference throughput by selecting the PvCs based on the attention of $S_r$. However, the process of selecting PvC introduces additional computation in each layer. As shown in Algorithm \ref{al:PyramidInfer}, the additional cost is mainly caused by the sort operation in top-$p$ while others can be neglected.

To evaluate the influence of the additional cost, we gradually increase the batch size of the models and compare the throughput between PyramidInfer and the full cache baseline. As shown in Figure \ref{fig:cost}, PyramidInfer has limited acceleration with a small batch size because the additional computation offsets the acceleration from the reduced KV cache. As the batch size increases, this cost becomes trivial compared to the acceleration brought by the PyramidInfer.

\subsection{Position Encoding}
As we reduce the number of keys and values of each layer, some positions of keys and values are missing. There are two choices to obtain the new position encoding: 1) re-encode the positions from position 0 in order; 2) gather the scattered original position encodings of the keys and values. As shown in Table \ref{tab:position}, we experiment on these two choices on LLaMA 2-13B and find that the latter one has a slightly better performance in the downstream tasks. 
\begin{table}[ht]
\centering
\setlength{\tabcolsep}{4.5pt}
\caption{Position encoding comparison.}
\label{tab:position}
 \small
{
\begin{tabular}{l|cc}
\toprule
{\textbf{Strategy}} & \textbf{GSM8K} & \textbf{MMLU} \\
\midrule
Re-encode & 29.12 & 55.5 \\
Gather & 29.56 & 55.7 \\
\bottomrule
\end{tabular}
}
\end{table}

\section{Extended Discussions}
\paragraph{The Association between ICR and RAC}
In Section \ref{sec:rac_discuss}, we mention the phenomenon that deeper layers have lower PvC overlap ratios is consistent with the power law distribution observed in Figure \ref{fig:cpr_std}. This is because, as we observe alone the layer index of the heatmap, we find that the color quickly deepens by a large gap where the depth change is approximate to the power law distribution.

The insight behind these two power law distributions is the same. The high redundancy in deeper layers indicates that most of the keys and values are useless for inference. These non-PvCs all have similarly low attention weights, resulting in limited influence on the perplexity and few opportunities to be selected as PvCs.

\begin{figure}[h]
    \centering
    \includegraphics[width=0.95\linewidth]{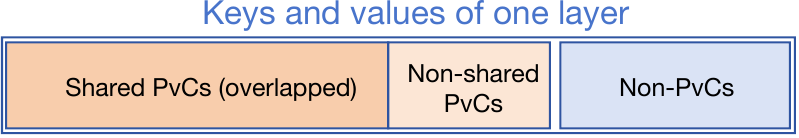}
    \caption{The composition of the keys and values of one layer.}
    \label{fig:pvc}
\end{figure}

\paragraph{Further Verification of ICR about the Role of Non-PvCs}
\label{app:icr}
To complete the verification of ICR, we have to verify the non-PvCs are redundant because they carry the information of predicting the tokens next to themselves instead of context information. In Figure \ref{fig:pvc}, to better illustrate, we divide the keys and values of one layer into two main parts, PvCs and non-PvCs. For the PvCs, we further divide them into shared PvCs and non-shared PvCs.

In Figure \ref{fig:rac_before}, we demonstrate that there is an 87\% overlap between tokens and the last token in terms of PvC, as denoted as shared PvC. We first identify the role of the remaining 13\% of keys and values where these non-shared PvCs are not used in PyramidInfer. The non-shared PvCs are also assigned high attention weights by the current token, which means they are useful for predicting the token next to the current token. It is interesting to see what these non-shared PvCs are from the perspective of the subsequent tokens: Will they also consider these keys and values important?

We use the recent sequence ratio of 20\% to select the shared PvCs. We extract non-shared PvCs from the tokens with $10\%<d<20\%$. We want to find these non-shared PvCs belong to which parts of keys and values of the subsequent tokens with $d<10\%$. 

\begin{figure*}
    \centering
    \includegraphics[width=0.9\linewidth]{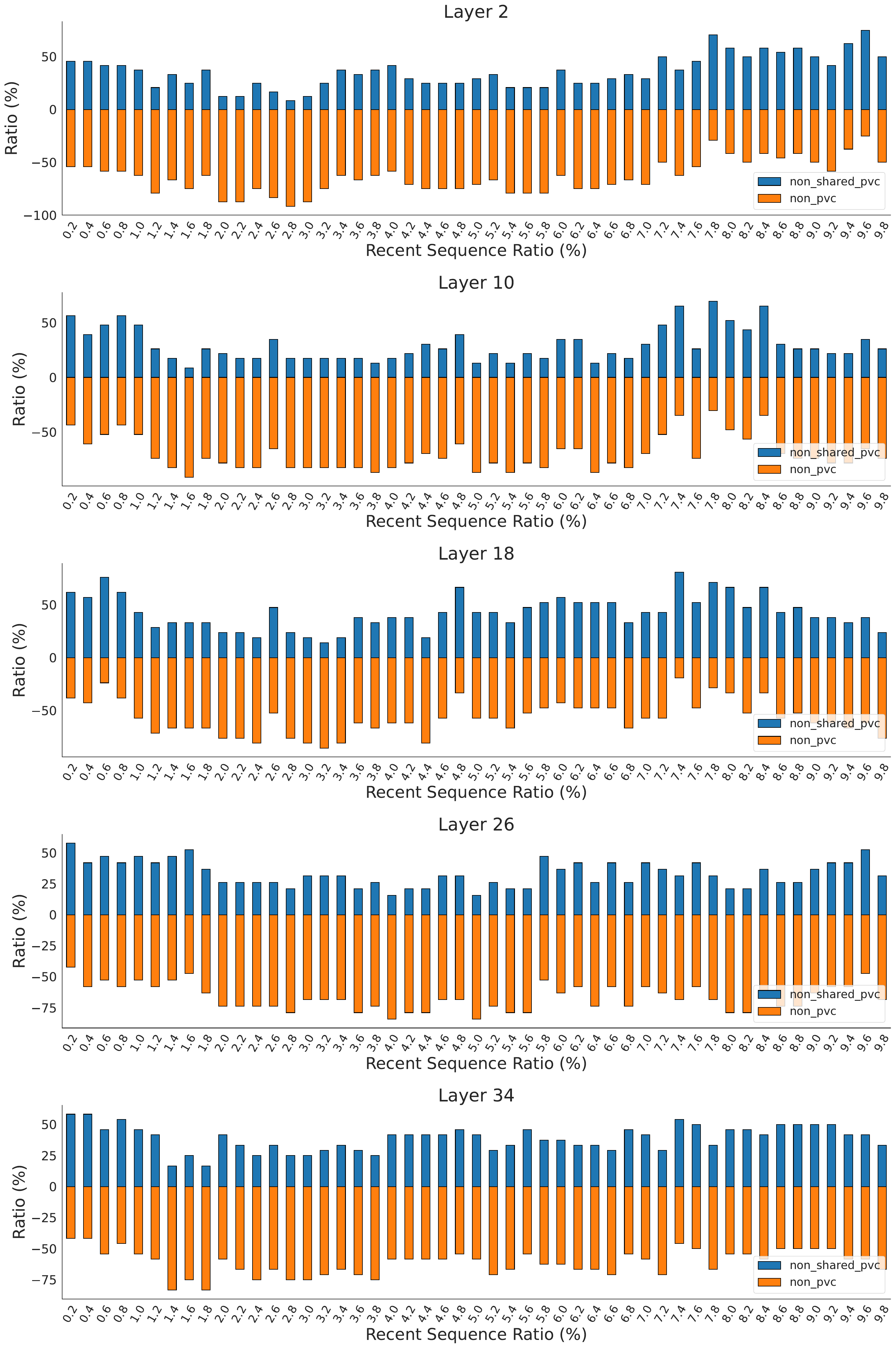}
    \caption{The overlap ratios between non-shared PvCs and non-shared PvCs of other tokens (blue) and the overlap ratios between non-shared PvCs and non-PvCs of other tokens (orange).}
    \label{fig:app_icr}
\end{figure*}

From Figure \ref{fig:app_icr}, we can draw conclusions for these three parts of the KV cache:
\begin{enumerate}
    \item The shared PvCs are the keys and values that subsequent tokens collectively pay attention to.
    \item The non-shared PvCs seldom appear in non-shared PvCs of other tokens. It means that non-shared PvCs are mostly highly interested in by the current token, with less attention from subsequent tokens. They are mainly used to predict the token next to themself in a teacher-forcing way, which is especially useful in training.
    \item Among the non-PvCs, a significant portion is occupied by non-shared PvCs of other tokens.
\end{enumerate}
So far, we have completely verified the Inference Context Redundancy hypothesis that the tokens except for the last token no longer need to predict the next tokens but they still record this redundant information to predict the next tokens in keys and values. 

\end{document}